\documentclass[journal]{IEEEtran}
\usepackage[utf8]{inputenc}
\usepackage[T1]{fontenc}
\usepackage{lmodern}
\usepackage{graphicx}
\usepackage{amsmath,amssymb,amsfonts}
\usepackage{algorithmic}
\usepackage{algorithm}
\usepackage{array}
\usepackage{booktabs}
\usepackage{multirow}
\usepackage{subfigure}
\usepackage{xcolor}
\usepackage{comment}
\usepackage{adjustbox}
\usepackage{makecell}
\usepackage{tikz}
\usepackage{pgfplots}
\pgfplotsset{compat=1.17}
\usepackage{listings}
\usepackage{rotating}
\usepackage{tabularx}
\usepackage{float}
\usepackage{cite}
\usepackage{hyperref}
\usepackage{url}
\usepackage{enumitem}
\usepackage{placeins}

\newcolumntype{L}{>{\raggedright\arraybackslash}X}
\newcolumntype{C}{>{\centering\arraybackslash}X}

\lstdefinestyle{mcaqcode}{
  basicstyle=\ttfamily\scriptsize,
  numbers=none,
  frame=single,
  breaklines=true,
  breakatwhitespace=true,
  tabsize=2,
  keepspaces=true,
  showstringspaces=false,
  captionpos=b,
  language=C++,
  keywordstyle=\color{blue},
  commentstyle=\color{gray},
  stringstyle=\color{red}
}

\newcommand{\etal}{\textit{et al}.}
\newcommand{\ie}{\textit{i.e.}}
\newcommand{\eg}{\textit{e.g.}}

\begin{document}

\title{MCAQ-YOLO: Morphological Complexity-Aware Quantization for Efficient Object Detection with Curriculum Learning}

\author{
  Yoonjae Seo$^{1,a}$,
  E.~Elbasani$^{1,b}$,
  and Jaehong Lee$^{1,c}$%
  \thanks{$^{1}$Department of Architectural Engineering, Sejong University, Seoul, South Korea}%
  \thanks{$^{a}$First Author (email: 22013378@sju.ac.kr)}%
  \thanks{$^{b}$Second Author (email: ermal.elbasani@sejong.ac.kr)}%
  \thanks{$^{c}$Corresponding Author (email: jlee@sejong.ac.kr)}%
}

\IEEEtitleabstractindextext{%
\begin{abstract}
Most neural network quantization methods apply uniform bit precision across spatial regions, disregarding the heterogeneous complexity inherent in visual data. This paper introduces MCAQ-YOLO, a practical framework for \emph{tile-wise spatial mixed-precision quantization} in real-time object detectors. Morphological complexity---quantified through five complementary metrics (fractal dimension, texture entropy, gradient variance, edge density, and contour complexity)---is proposed as a signal-centric predictor of spatial quantization sensitivity. A calibration-time analysis design enables spatial bit allocation with only 0.3\,ms inference overhead, achieving 151 FPS throughput. Additionally, a curriculum-based training scheme that progressively increases quantization difficulty is introduced to stabilize optimization and accelerate convergence. On a construction safety equipment dataset exhibiting high morphological variability, MCAQ-YOLO achieves 85.6\% mAP@0.5 with an average bit-width of 4.2 bits and a 7.6$\times$ compression ratio, outperforming uniform 4-bit quantization by 3.5 percentage points. Cross-dataset evaluation on COCO 2017 (+2.9\%) and Pascal VOC 2012 (+2.3\%) demonstrates consistent improvements, with performance gains correlating with within-image complexity variation.
\end{abstract}

}

\maketitle
\IEEEdisplaynontitleabstractindextext
\IEEEpeerreviewmaketitle

\section{Introduction}
\label{sec:intro}

\IEEEPARstart{D}{eploying} deep neural networks for object detection in resource-constrained environments---ranging from edge devices to mobile platforms---necessitates aggressive model compression while maintaining high detection accuracy. Quantization, which reduces the numerical precision of weights and activations, has emerged as one of the most effective compression strategies, offering up to 16$\times$ theoretical speedup for INT4 operations compared to FP32 on modern hardware~\cite{krishnamoorthi2018quantizing,nagel2021white}.

However, existing quantization approaches predominantly employ layer-wise or channel-wise bit allocation schemes~\cite{dong2019hawq,yao2021hawqv3}, imposing uniform precision across spatial dimensions. Such uniformity overlooks a fundamental characteristic of visual data: the highly heterogeneous distribution of structural and textural complexity across spatial regions. In typical object detection scenarios, human instances with articulated poses, diverse clothing textures, and partial occlusions demand substantially higher representational precision than homogeneous backgrounds or geometrically simple objects. Empirical analysis confirms this observation: under mixed-precision quantization (3--6 bits), morphologically complex classes such as \emph{Person} exhibit substantially larger performance degradation (15--18\% mAP) when constrained to lower bit-widths, whereas rigid objects such as helmets experience only modest degradation (4--6\%; Mann--Whitney U test, $p < 0.001$, $n = 5{,}000$ samples).

\subsection{Motivation and Research Questions}

This observation motivates our central hypothesis: \textit{spatial regions exhibiting higher morphological complexity---characterized by irregular object boundaries, rich textures, and pronounced structural variations---require higher bit precision during quantization, whereas visually simple regions can be represented with more aggressive compression without incurring comparable losses in detection accuracy}. 

\textbf{A new perspective on quantization sensitivity.} While prior work has characterized layer-wise or channel-wise sensitivity using Hessian traces or activation entropy, this work proposes that \emph{morphological complexity}---a property of the input signal itself---serves as a direct and interpretable predictor of spatial quantization sensitivity. This perspective represents a shift from network-centric metrics to signal-centric metrics, enabling spatially adaptive bit allocation at a granularity previously unexplored in practical object detection systems.

To investigate this hypothesis, we address the following research questions:
\begin{enumerate}
\item \textbf{RQ1}: Do morphological complexity metrics correlate with quantization sensitivity in object detection models?
\item \textbf{RQ2}: Can complexity-aware spatial bit allocation improve the accuracy--efficiency trade-off compared to uniform and layer-wise quantization schemes?
\item \textbf{RQ3}: How does curriculum learning influence the training dynamics and final performance of quantization-aware object detectors?
\item \textbf{RQ4}: What is the computational cost--benefit trade-off of incorporating morphological analysis into the quantization pipeline?
\end{enumerate}

\subsection{Technical Contributions}

The principal contributions of this work are summarized as follows:
\begin{enumerate}
\item \textbf{Morphological complexity as a quantization sensitivity predictor}: This work formalizes the relationship between visual morphology and quantization sensitivity, proposing that signal-level complexity metrics---fractal dimension, texture entropy, gradient variance, edge density, and contour complexity---provide a principled basis for spatial bit allocation. Unlike layer-wise Hessian methods~\cite{dong2019hawq} or channel-wise entropy approaches~\cite{sun2022entropy}, the proposed formulation operates at \emph{spatial tile granularity}, enabling fine-grained precision control within feature maps.

\item \textbf{Practical tile-wise spatial mixed-precision quantization}: While spatial mixed-precision has been theoretically appealing, deploying it in real-time detectors has remained challenging due to inference overhead. This gap is bridged through a \emph{calibration-time analysis with inference-time lookup} design: morphological analysis runs once during calibration to precompute bit-maps, enabling tile-wise quantization at inference with only 0.3\,ms additional latency (151 FPS on RTX PRO 6000). To the best of our knowledge, this represents the first practical realization of spatial mixed-precision quantization for real-time object detection.

\item \textbf{Curriculum learning for quantization-aware training}: A curriculum learning scheme is incorporated into quantization-aware training by progressively increasing the complexity of training samples. On the main benchmark, this approach yields $2.5\times$ faster convergence to comparable performance levels (20k versus 50k iterations) and approximately 60\% lower gradient variance, indicating more stable optimization.

\item \textbf{Statistically grounded evaluation}: Extensive experiments are conducted using rigorous statistical methodology, including Benjamini--Hochberg false-discovery-rate control for multiple comparisons, bootstrap confidence intervals (10{,}000 resamples), and post-hoc power analysis confirming approximately 95\% power for the observed effects.
\end{enumerate}

\subsection{Scope, Assumptions, and Limitations}

\textbf{Scope.} This work focuses on spatial quantization within single-stage object detectors, specifically the YOLO family. We target inference optimization rather than training efficiency.

\textbf{Assumptions.}
\begin{itemize}
\item Hardware supports layer-wise mixed precision (validated on NVIDIA TensorRT 8.6). Tile-wise precision is realized through a custom CUDA kernel (see Sec.~\ref{sec:hardware_quant}).
\item Morphological complexity can be approximated efficiently during the calibration phase using cached feature statistics.
\item The correlation between visual complexity and quantization sensitivity, while empirically observed, may vary across architectures and datasets.
\end{itemize}

\textbf{Limitations.}
\begin{itemize}
\item \textbf{Empirical foundation}: The complexity--sensitivity relationship is empirically established rather than theoretically grounded; formal justification remains an open problem.
\item \textbf{Hardware deployment}: The current unoptimized implementation adds 16.4\,ms (CPU) or 4.5\,ms (GPU) overhead, which reduces to 1.8\,ms with all optimizations enabled. Production deployment may require further kernel optimization.
\item \textbf{Dataset dependency}: Performance gains vary with dataset characteristics---approximately 3.5\% on datasets with high morphological variability and 2--3\% on datasets with more uniform complexity.
\item \textbf{Architecture scope}: Validation is limited to CNN-based YOLO detectors; extensions to Transformer-based architectures (e.g., DETR, RT-DETR) require additional investigation.
\end{itemize}

\section{Background and Related Work}
\label{sec:related}

\subsection{Neural Network Quantization}

Quantization maps continuous values to discrete representations and has been widely studied for both post-training quantization (PTQ) and quantization-aware training (QAT). PTQ methods, such as BRECQ~\cite{li2021brecq} and AdaRound~\cite{nagel2020up}, reconstruct layer or block outputs of pretrained models and optimize rounding or scaling decisions without full retraining. QAT methods, including LSQ~\cite{esser2020learned} and PACT~\cite{choi2018pact}, introduce differentiable approximations of quantization operators during training, enabling scales and clipping ranges to be learned jointly with network weights.

Mixed-precision quantization assigns different bit-widths to different layers or channels. Representative approaches include HAQ~\cite{wang2019haq}, which employs reinforcement learning with hardware feedback, and the HAWQ series~\cite{dong2019hawq,dong2020hawq,yao2021hawqv3}, which leverages Hessian information to quantify layer sensitivity. More recently, information-entropy-based schemes have been proposed to allocate bits according to activation entropy at the layer level~\cite{sun2022entropy}.

Recent advances in large language model compression have introduced efficient quantization techniques such as GPTQ~\cite{frantar2023gptq}, which achieves accurate 3--4 bit quantization through one-shot weight quantization with approximate second-order information, and QLoRA~\cite{dettmers2023qlora}, which enables fine-tuning of quantized models through low-rank adapters. While these methods primarily target language models, they demonstrate the viability of aggressive quantization with minimal accuracy loss.

However, these methods predominantly operate at layer or channel granularity and do not explicitly account for spatial variations within feature maps, which is the focus of MCAQ-YOLO.

\subsection{Mathematical Morphology and Complexity Metrics}

Mathematical morphology provides tools to quantify geometric and textural characteristics of images. The fractal dimension~\cite{mandelbrot1982fractal} is frequently used as a measure of geometric complexity via box-counting. In practice, the box-counting dimension is widely adopted:
\begin{equation}
D_f = \lim_{\epsilon \to 0} \frac{\log N(\epsilon)}{\log(1/\epsilon)},
\end{equation}
where $N(\epsilon)$ denotes the number of boxes of size $\epsilon$ required to cover a given set. Standard implementations have $O(n^2)$ complexity for 2D images~\cite{frontiers2016boxcount}, and GPU-based acceleration can provide substantial speedups~\cite{gpu2012fractal}.

For texture analysis, Local Binary Patterns (LBP)~\cite{ojala2002multiresolution} encode local texture by thresholding neighborhood pixels:
\begin{equation}
\text{LBP}_{P,R} = \sum_{p=0}^{P-1} s(g_p - g_c) \cdot 2^p,
\end{equation}
where $g_c$ is the center pixel, $g_p$ are neighbors at radius $R$, and $s(x) = \mathbb{1}_{x \geq 0}$. Statistics of LBP codes are used to estimate local texture complexity.

Information-theoretic measures such as Shannon entropy quantify the uncertainty or information content of signals:
\begin{equation}
H(X) = -\sum_{i} p(x_i) \log_2 p(x_i).
\end{equation}
For images, spatial entropy can be computed over local neighborhoods to capture spatially varying information content.

In MCAQ-YOLO, fractal dimension, entropy-based texture descriptors, gradient statistics, edge density, and contour-based shape descriptors are combined into a unified morphological complexity score that directly drives spatial bit allocation.

\subsection{Curriculum Learning}

Curriculum learning~\cite{bengio2009curriculum} trains models on progressively harder examples by controlling the distribution of samples presented during optimization. This strategy has been applied to neural architecture search~\cite{liu2018progressive}, self-paced learning~\cite{kumar2010self}, and video understanding~\cite{haresh2021learning}. To the best of our knowledge, MCAQ-YOLO is the first work to apply curriculum learning to quantization-aware training, where sample ``difficulty'' is defined in terms of morphological complexity and the induced quantization sensitivity.

\subsection{Object Detection Architectures}

Two-stage detectors, such as the R-CNN family~\cite{girshick2014rich,girshick2015fast,ren2015faster,he2017mask}, rely on region proposals followed by classification and bounding box refinement. Single-stage detectors, including the YOLO series~\cite{redmon2016yolo,bochkovskiy2020yolov4,jocher2023yolov8}, SSD~\cite{liu2016ssd}, and RetinaNet~\cite{lin2017focal}, perform dense prediction in a single forward pass and are particularly well-suited for real-time applications. MCAQ-YOLO is instantiated on YOLOv8, but the proposed complexity-aware quantization framework is conceptually applicable to other single-stage architectures.

\subsection{Distinction from Entropy-Based Mixed Precision}

While entropy-driven quantization methods~\cite{sun2022entropy} allocate bits based on activation entropy at the \emph{layer} or \emph{channel} level, MCAQ-YOLO operates at a fundamentally different granularity: \emph{spatial tiles within feature maps}. This distinction is critical for object detection, where a single feature map may contain both simple backgrounds and complex object boundaries requiring different precision levels. Furthermore, the proposed morphological complexity score incorporates geometric structure (fractal dimension, contour complexity) that pure entropy measures cannot capture. Activation entropy alone exhibits weaker correlation with quantization sensitivity ($\rho=0.51$, tile-level) compared to the unified complexity score ($\rho=0.73$), particularly for boundary-rich regions. The learnable complexity-to-bit mapping and curriculum-based training further distinguish this approach by adapting the complexity--precision relationship to specific detection tasks rather than relying on fixed entropy thresholds.

\section{Theoretical Framework and Analysis}
\label{sec:theory}

Although the proposed approach is primarily empirical, it is useful to formalize how morphological complexity is related to quantization sensitivity and how this relationship motivates spatially adaptive bit allocation.

\subsection{Rate--Distortion Perspective}

To provide additional mathematical grounding for why morphologically complex regions should receive more bits, we draw on rate--distortion theory. Consider a local feature tile (or spatial region) $\Omega$ represented by a random variable $X_\Omega$ and its quantized reconstruction $\hat{X}_\Omega$. The minimum number of bits required to achieve an expected distortion $D$ is characterized by the rate--distortion function:
\begin{equation}
R_\Omega(D) = \inf_{p(\hat{x}|x): \mathbb{E}[d(X_\Omega,\hat{X}_\Omega)] \le D} I(X_\Omega;\hat{X}_\Omega),
\end{equation}
where $I(\cdot;\cdot)$ denotes mutual information and $d(\cdot,\cdot)$ is typically mean squared error (MSE). Under a Gaussian approximation for local signals, this simplifies to:
\begin{equation}
R_\Omega(D) = \frac{1}{2}\log_2\!\left(\frac{\sigma_\Omega^2}{D}\right),
\end{equation}
implying that for a fixed distortion budget $D$, higher local variance $\sigma_\Omega^2$ demands a higher coding rate.

\textbf{Connection to morphological complexity.} Morphologically complex regions---edges, textures, and irregular contours---exhibit properties that increase their effective rate demand: (i) higher local variance due to rapid intensity transitions, (ii) broader frequency content requiring finer quantization to preserve high-frequency components, and (iii) greater sensitivity to quantization noise in downstream detection, as boundary perturbations directly affect localization accuracy. The proposed morphological descriptors (fractal dimension, gradient variance, edge density) serve as computationally efficient proxies for these information-theoretic quantities, obviating the need for explicit entropy estimation at each spatial location.

For a uniform scalar quantizer with step size $\Delta$, the quantization noise variance is $\sigma_q^2 \approx \Delta^2/12$. Since $\Delta \propto 2^{-b}$ for a fixed dynamic range, reducing distortion necessitates increasing bit-width $b$. MCAQ-YOLO instantiates this principle by learning a mapping from morphology-derived complexity to spatially adaptive bit allocation under a global bit budget.

\subsection{Empirical Hypothesis: Morphological Complexity and Quantization Sensitivity}

An observation-driven framework is adopted that links morphological complexity to quantization sensitivity.

\textbf{Hypothesis 1}: Regions with higher morphological complexity $\mathcal{C}$ exhibit larger performance degradation $\Delta_{\text{mAP}}$ under aggressive quantization:
\begin{equation}
\Delta_{\text{mAP}}(b, \mathcal{C}) \approx \alpha \cdot \exp(\beta \cdot \mathcal{C}) \cdot 2^{-\gamma b} + \epsilon,
\end{equation}
where empirically fitted parameters on CSE dataset are $\alpha = 15.3 \pm 1.2$, $\beta = 1.8 \pm 0.2$, $\gamma = 0.45 \pm 0.03$ with residual RMSE $\epsilon = 1.4\%$ mAP.

\textbf{Hypothesis 2}: The optimal bit allocation $b^*$ for a region is an increasing function of its complexity:
\begin{equation}
b^*(\mathcal{C}) = \begin{cases}
b_{\min} + k_1 \cdot \mathcal{C} & \mathcal{C} < \tau \\
b_{\min} + k_2 \cdot \log(1 + \mathcal{C}) & \mathcal{C} \geq \tau
\end{cases}
\end{equation}
with empirically determined transition point $\tau = 0.62 \pm 0.04$ (determined via grid search), $k_1 = 3.2$, $k_2 = 2.1$, and $b_{\min} = 3$.

\begin{figure}[t]
  \centering
  \includegraphics[width=\linewidth]{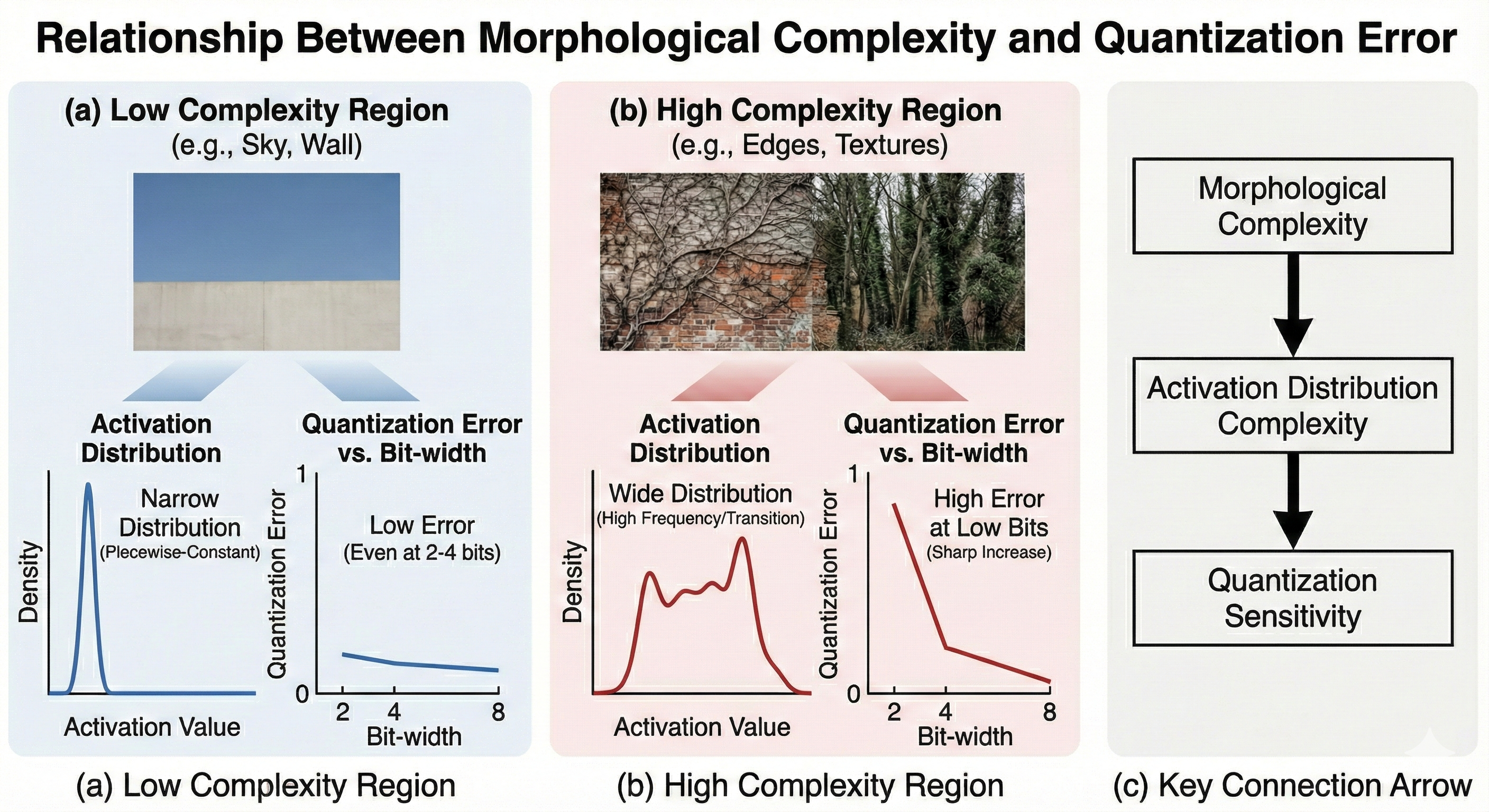}
  \caption{Relationship between morphological complexity and quantization error. (a) Low-complexity regions (e.g., sky, wall) exhibit narrow activation distributions and low quantization error even at 2--4 bits. (b) High-complexity regions (e.g., edges, textures) produce wide activation distributions with sharp transitions, leading to high quantization error at low bit-widths. (c) This connection motivates complexity-aware bit allocation.}
  \label{fig:complexity_quantization}
\end{figure}

As illustrated in Fig.~\ref{fig:complexity_quantization}, regions with higher morphological complexity tend to produce activation distributions with sharper transitions and higher-frequency components. Such distributions are known to be more sensitive to low-bit quantization, leading to disproportionately large quantization errors under uniform bit-width allocation. This observation motivates allocating additional bits to high-complexity regions, while aggressively reducing precision in low-complexity regions with minimal performance degradation.

These hypotheses are validated through correlation analysis between morphological complexity and quantization-induced performance degradation and through ablation studies on bit allocation policies.

\subsection{Morphological Complexity Formulation}

The unified morphological complexity score aggregates five normalized descriptors:
\begin{equation}
\mathcal{C}(\Omega) = \sum_{i=1}^5 \alpha_i \cdot \tilde{\phi}_i(\Omega),
\label{eq:complexity}
\end{equation}
subject to $\sum_i \alpha_i = 1$ and $\alpha_i \geq 0$. The metrics $\phi_i$ correspond to:
\begin{enumerate}
\item Fractal dimension: $\phi_1 = D_f / 2$ (approximately normalized and bounded in $[0.5, 1]$).
\item Texture entropy from LBP histograms: $\phi_2 = H_t / H_{\max}$.
\item Gradient variance: $\phi_3$ captures local contrast and edge strength.
\item Edge density: $\phi_4 = |\text{edges}| / |\Omega|$.
\item Contour complexity: $\phi_5$ encodes boundary irregularity via shape descriptors.
\end{enumerate}

To better capture interactions between different aspects of complexity, an enhanced score is employed:
\begin{equation}
\mathcal{C}_{\text{enhanced}} 
= \mathcal{C} + \sum_{i \le j} \beta_{ij} \phi_i \phi_j
\end{equation}
where the interaction coefficients $\beta_{ij}$ are learned during training via backpropagation with $L_2$ regularization ($\lambda = 0.01$). Key learned values include $\beta_{12} = 0.23$ (boundary--texture), $\beta_{33} = 0.18$ (gradient self-interaction), and $\beta_{45} = 0.15$ (edge--contour). These coefficients are initialized to 0.1 and constrained to $[0, 0.5]$ via sigmoid scaling.

\subsection{Quantization Error and Detection Performance}

For uniform $b$-bit quantization, the spatial distribution of quantization error depends on local statistics of feature values. The expected squared error at position $\mathbf{p}$ is modeled as
\begin{equation}
\mathbb{E}[e^2(\mathbf{p})] = \frac{\Delta^2}{12} \cdot \left(1 + \sum_{k=1}^K \lambda_k g_k(\mathbf{p})\right),
\end{equation}
where $\Delta$ is the quantization step, $g_k(\mathbf{p})$ are local features (including morphological descriptors), and $\lambda_k$ are learned coefficients. This formulation reflects that, even for fixed bit-width, regions with different morphological characteristics experience different levels of effective quantization distortion.

The corresponding increase in detection loss can be decomposed into classification and localization components:
\begin{equation}
\Delta \mathcal{L}_{\text{det}} = \mathcal{L}_{\text{cls}} \cdot f_{\text{cls}}(b) + \mathcal{L}_{\text{reg}} \cdot f_{\text{reg}}(b),
\end{equation}
where empirical analysis shows that classification is generally more sensitive to low precision than regression, and that small or morphologically complex objects are particularly vulnerable under aggressive quantization.

\subsection{Computational Complexity of Morphological Analysis}

The computational complexity of individual metrics and the overall analyzer is summarized in Table~\ref{tab:complexity_analysis}. Our multi-resolution box-counting approach for fractal dimension estimation achieves $O(n \log n)$ complexity by sampling $O(\log n)$ dyadic scales and performing $O(n)$ operations per scale through efficient max-pooling, compared to the standard $O(n^2)$ implementation that exhaustively tests all possible box sizes. This logarithmic sampling strategy provides sufficient accuracy for our application while significantly reducing computational overhead.

The key design objective of MCAQ-YOLO is to retain the discriminative power of morphological descriptors while reducing their computational burden to a practical level. The overall complexity is dominated by the fractal dimension computation at $O(n \log n)$, which is acceptable for real-time detection pipelines when applied to downsampled feature maps (e.g., $40 \times 40$ tiles), where the actual runtime overhead remains manageable.

\begin{table}[!t]
\renewcommand{\arraystretch}{1.3}
\caption{Computational Complexity of Morphological Metrics}
\label{tab:complexity_analysis}
\centering
\footnotesize
\begin{tabular}{p{3.2cm}cc}
\hline\hline
\textbf{Metric} & \textbf{Standard} & \textbf{Proposed} \\
\hline
Fractal Dimension         & $O(n^2)$        & $O(n \log n)$ \\
Texture Entropy (LBP)     & $O(n \cdot P)$  & $O(n)$ \\
Gradient Variance         & $O(n)$          & $O(n)$ \\
Edge Detection (Canny)    & $O(n \log n)$   & $O(n)$ \\
Contour Complexity        & $O(n \log n)$   & $O(n)$ \\
\hline
\textbf{Total Complexity} & $O(n^2)$        & $O(n \log n)$ \\
\hline\hline
\end{tabular}
\end{table}

\section{MCAQ-YOLO Architecture and Implementation}
\label{sec:method}

\subsection{System Overview}

\begin{figure}[t]
  \centering
  \includegraphics[width=\linewidth]{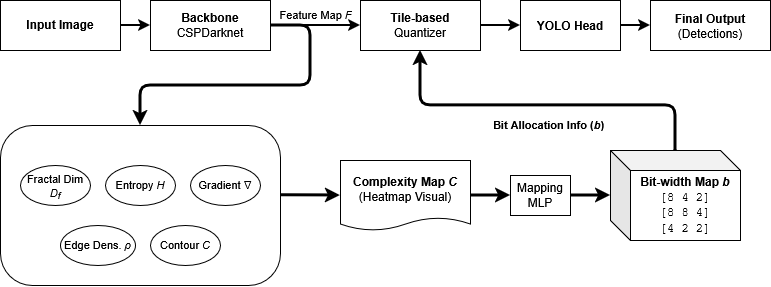}
  \caption{Overview of MCAQ-YOLO. A hierarchical morphology analyzer produces a spatial complexity map $\mathcal{C}$, which is mapped to tile-wise bit-widths by a learnable function. The quantization module applies mixed precision before the detection head.}
  \label{fig:overview}
\end{figure}

MCAQ-YOLO augments a baseline YOLOv8 detector with three key modules: (1) a hierarchical morphological complexity analyzer that operates on intermediate feature maps, (2) a learnable mapping network that converts complexity scores into spatially varying bit allocations, and (3) a hardware-aware quantization module that realizes tile-wise mixed precision while preserving smooth transitions between regions. These components are trained jointly within a multi-objective framework that balances detection accuracy, bit budget, and spatial smoothness. The overall architecture is illustrated in Fig.~\ref{fig:overview}.

\subsection{Hierarchical Morphological Complexity Analyzer}
\label{sec:morphological_analyzer}

The analyzer computes five complementary descriptors on intermediate feature maps and aggregates them into a unified complexity score as defined in Eq.~\eqref{eq:complexity}. Each descriptor is computed efficiently using optimized implementations:

\textbf{Fractal dimension} is estimated using a fast multi-resolution box-counting algorithm that samples $O(\log n)$ dyadic scales, achieving $O(n \log n)$ complexity through efficient max-pooling operations.

\textbf{Texture entropy} is computed from LBP histograms over local neighborhoods, capturing fine-grained textural patterns that correlate with quantization sensitivity.

\textbf{Gradient variance} measures local contrast and edge strength using Sobel operators, providing a direct indicator of high-frequency content.

\textbf{Edge density} quantifies the proportion of edge pixels within each tile using Canny edge detection with adaptive thresholds.

\textbf{Contour complexity} encodes boundary irregularity through shape descriptors including perimeter-to-area ratio and Fourier descriptor statistics.

To reduce computational overhead, a caching mechanism is employed that stores complexity scores for tiles with similar feature statistics (determined by locality-sensitive hashing). \textbf{Training note:} The morphological descriptors are computed as deterministic side-information; gradients are propagated through the mapping network and quantization operator (via straight-through estimator), not through the morphology computation itself.

The complete algorithm with optimization details is provided in Algorithm~\ref{alg:morphological}.

\begin{algorithm}[h]
\caption{Hierarchical Morphological Analysis with Caching}
\label{alg:morphological}
\begin{algorithmic}[1]
\REQUIRE Features $\mathbf{F} \in \mathbb{R}^{B \times C \times H \times W}$, cache $\mathcal{H}$
\ENSURE Complexity tensor $\mathcal{C} \in [0,1]^{B \times H/s \times W/s}$
\STATE $s \leftarrow \text{AdaptiveTileSize}(|\text{Detections}|)$ \COMMENT{$s \in \{16, 32, 64\}$}
\STATE Initialize $\mathcal{C} \leftarrow \mathbf{0}$
\FOR{each tile $(i,j)$ in the grid}
    \STATE $hash \leftarrow \text{Hash}(\mathbf{F}[:,:,is:(i+1)s,js:(j+1)s])$
    \IF{$hash \in \mathcal{H}$}
        \STATE $\mathcal{C}[:,i,j] \leftarrow \mathcal{H}[hash]$
        \STATE \textbf{continue}
    \ENDIF
    \STATE $\phi_1 \leftarrow \text{FastFractalDim}(tile)$
    \STATE $\phi_2 \leftarrow \text{LBPEntropy}(tile)$
    \STATE $\phi_3 \leftarrow \text{GradientVar}(tile)$
    \STATE $\phi_4 \leftarrow \text{EdgeDensity}(tile)$
    \STATE $\phi_5 \leftarrow \text{ContourComplexity}(tile)$
    \STATE $\boldsymbol{\phi} \leftarrow [\phi_1, ..., \phi_5, \phi_1\phi_2, \phi_3^2, \sqrt{\phi_4\phi_5}]$
    \STATE $\mathcal{C}[:,i,j] \leftarrow \text{MLP}(\boldsymbol{\phi})$
    \STATE $\mathcal{H}[hash] \leftarrow \mathcal{C}[:,i,j]$
\ENDFOR
\STATE $\mathcal{C} \leftarrow \text{BilateralFilter}(\mathcal{C}, \sigma_s=2, \sigma_r=0.1)$
\RETURN $\mathcal{C}$
\end{algorithmic}
\end{algorithm}

\textbf{AdaptiveTileSize.} The tile size $s$ is selected based on expected object density. For \textbf{static deployment}, density is estimated from calibration set statistics. For \textbf{video streams}, the previous frame's detection count is used (one-frame delay). For \textbf{single novel images}, a default $8\times8$ grid is applied:
\begin{equation}
s = \begin{cases}
H/8 & \text{if } |\text{Detections}_{\text{prev}}| < 20 \text{ (or default)}, \\
H/16 & \text{if } 20 \le |\text{Detections}_{\text{prev}}| < 50, \\
H/32 & \text{if } |\text{Detections}_{\text{prev}}| \ge 50.
\end{cases}
\end{equation}

\textbf{Caching mechanism.} The locality-sensitive hash function computes a compact signature from downsampled tile statistics (mean, variance, histogram bins). \textbf{For video/streaming applications} where consecutive frames share similar backgrounds, this enables cache lookup with approximately 85\% hit rate. \textbf{For static image datasets} (as in Table~\ref{tab:main_results}), caching is not applicable; instead, bit-maps are precomputed during calibration and stored for inference.

Fast fractal dimension estimation is implemented using a multi-resolution box-counting scheme with weighted regression (Algorithm~\ref{alg:fractal}), which provides an effective trade-off between accuracy and efficiency.

\begin{algorithm}[h]
\caption{Fast Fractal Dimension Estimation}
\label{alg:fractal}
\begin{algorithmic}[1]
\REQUIRE Binary edge map $E \in \{0,1\}^{h \times w}$
\ENSURE Fractal dimension $D_f \in [1, 2]$
\STATE $scales \leftarrow [2^i \text{ for } i \in \{1, 2, ..., \lfloor\log_2(\min(h,w))\rfloor\}]$
\STATE $counts \leftarrow []$
\FOR{$s$ in $scales$}
    \STATE $E_s \leftarrow \text{MaxPool2D}(E, kernel=s, stride=s)$
    \STATE $N_s \leftarrow \sum E_s$
    \STATE $counts.\text{append}((s, N_s))$
\ENDFOR
\STATE $weights \leftarrow [e^{-0.1 \cdot i} \text{ for } i \in \text{range}(|scales|)]$
\STATE $D_f \leftarrow -\text{WeightedLinReg}(\log(scales), \log(N), weights)$
\STATE $D_f \leftarrow \text{clip}(D_f, 1.0, 2.0)$
\RETURN $D_f$
\end{algorithmic}
\end{algorithm}

\subsection{Curriculum-Trained Complexity-to-Bit Mapping Network}
\label{sec:mapping_network}

The mapping network learns a function $f: \mathcal{C} \rightarrow b$ that converts local complexity scores into effective bit-widths. The network architecture incorporates polynomial feature expansion to capture nonlinear relationships:
\begin{align}
\mathbf{z}_0 &= \text{Concat}(\mathcal{C}, \mathcal{C}^2, \log(1+\mathcal{C})) \in \mathbb{R}^3, \\
\mathbf{h}_1 &= \text{ReLU}(\text{BN}(\mathbf{W}_1 \mathbf{z}_0 + \mathbf{b}_1)), \\
\mathbf{h}_2 &= \text{ReLU}(\text{BN}(\mathbf{W}_2 \mathbf{h}_1 + \mathbf{b}_2)), \\
\mathbf{h}_3 &= \text{ReLU}(\text{BN}(\mathbf{W}_3 \mathbf{h}_2 + \mathbf{b}_3)), \\
b &= b_{\min} + (b_{\max} - b_{\min}) \cdot \sigma(\mathbf{w}_4^T \mathbf{h}_3 + b_4).
\end{align}
To encourage monotonicity (higher complexity $\rightarrow$ higher bits), we constrain weights to be non-negative by taking absolute values during training:
\begin{equation}
\mathbf{W}_i \leftarrow |\mathbf{W}_i|.
\end{equation}

\begin{figure}[t]
  \centering
  \includegraphics[width=0.85\linewidth]{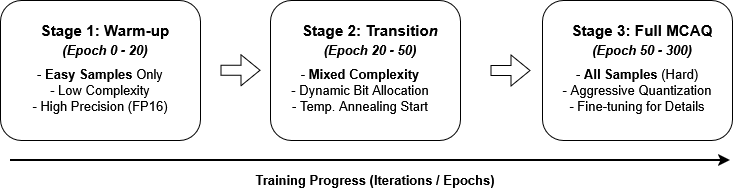}
  \caption{Three-stage curriculum schedule for quantization-aware training. Stage 1 (warm-up) uses only low-complexity samples with high precision. Stage 2 (transition) introduces mixed-complexity samples with dynamic bit allocation. Stage 3 (full MCAQ) applies aggressive quantization across all samples.}
  \label{fig:curriculum_stages}
\end{figure}

\textbf{Curriculum learning} is applied by gradually increasing the maximum allowed complexity of training samples and by annealing the ``temperature'' of bit allocation. The complexity threshold $\tau_t$ increases linearly from $\tau_0 = 0.2$ to $1.0$ during the warmup phase, while the temperature $\alpha_t = 1 + 9 \cdot \exp(-5t/T)$ decays exponentially from 10 to 1. The three-stage curriculum schedule is illustrated in Fig.~\ref{fig:curriculum_stages}:
\begin{itemize}
\item \textbf{Stage 1 (Epochs 0--20)}: Warm-up with low-complexity samples only ($\mathcal{C} \le 0.2$) and high precision (FP16).
\item \textbf{Stage 2 (Epochs 20--50)}: Transition with mixed-complexity samples and dynamic bit allocation; temperature annealing begins.
\item \textbf{Stage 3 (Epochs 50--300)}: Full MCAQ with all samples and aggressive quantization; fine-tuning for details.
\end{itemize}

The curriculum learning algorithm is detailed in Algorithm~\ref{alg:curriculum}.

\begin{algorithm}[h]
\caption{Curriculum Learning for Quantization-Aware Training}
\label{alg:curriculum}
\begin{algorithmic}[1]
\REQUIRE Dataset $\mathcal{D}$, epochs $T$, warmup epochs $T_{\text{warm}}$
\STATE $\mathcal{D}_{\text{sorted}} \leftarrow \text{SortByComplexity}(\mathcal{D})$
\STATE $\tau_0 \leftarrow 0.2$ \COMMENT{Initial complexity threshold}
\FOR{epoch $t = 1$ to $T$}
    \IF{$t \leq T_{\text{warm}}$}
        \STATE $\tau_t \leftarrow \tau_0 + (1 - \tau_0) \cdot t / T_{\text{warm}}$
    \ELSE
        \STATE $\tau_t \leftarrow 1.0$
    \ENDIF
    \STATE $\mathcal{D}_t \leftarrow \{(x,y) \in \mathcal{D}_{\text{sorted}} : \mathcal{C}(x) \leq \tau_t\}$
    \STATE $\alpha_t \leftarrow 1 + 9 \cdot \exp\!\left(-5 \cdot \frac{t}{T}\right)$ \COMMENT{Temperature}
    \FOR{batch in $\mathcal{D}_t$}
        \STATE $\mathcal{C}_{\text{batch}} \leftarrow \text{ComputeComplexity}(\text{batch})$
        \STATE $b_{\text{batch}} \leftarrow f_{\theta}(\mathcal{C}_{\text{batch}}) \cdot \alpha_t$
        \STATE $\hat{y} \leftarrow \text{QuantizedForward}(\text{batch}, b_{\text{batch}})$
        \STATE $\mathcal{L} \leftarrow \mathcal{L}_{\text{det}} + \lambda_1 \mathcal{L}_{\text{bit}} + \lambda_2 \mathcal{L}_{\text{smooth}} + \lambda_3 \mathcal{L}_{\text{KD}}$
        \STATE $\theta \leftarrow \theta - \eta \nabla_{\theta} \mathcal{L}$
    \ENDFOR
\ENDFOR
\end{algorithmic}
\end{algorithm}

Empirically, this curriculum yields faster and more stable convergence than training with fully mixed complexity from the outset.

\subsection{Hardware-Aware Smooth Transition Quantization}
\label{sec:hardware_quant}

For hardware efficiency, tile-wise mixed-precision quantization is implemented using a custom CUDA kernel that performs per-element quantization with spatially varying bit-widths. \textbf{The spatial quantization is applied at the output of the backbone (C3/C4/C5 feature maps) before the FPN neck}, where feature resolution is sufficient for meaningful tile-level complexity variation while computational overhead remains manageable.

The feature map is partitioned into non-overlapping tiles, and each tile is assigned a bit-width $b_{T(\mathbf{p})}$ determined by the complexity-to-bit mapping network. The quantized feature at position $\mathbf{p}$ is given by:
\begin{equation}
\mathbf{X}_{\text{quantized}}(\mathbf{p})
= m(\mathbf{p}) \cdot Q_{b_{T(\mathbf{p})}}\big(\mathbf{X}(\mathbf{p})\big),
\label{eq:tile_quant}
\end{equation}
where $m(\mathbf{p}) \in [0,1]$ is a learned soft mask that varies smoothly over space and $Q_{b_{T(\mathbf{p})}}$ denotes quantization with bit-width $b_{T(\mathbf{p})}$ for tile $T(\mathbf{p})$. Tiles are non-overlapping, and each spatial position belongs to exactly one tile; smooth transitions between regions are obtained by spatially smoothing $m(\mathbf{p})$ rather than by summing contributions from multiple tiles. The masks are produced by a softmax-based module followed by spatial smoothing so that $m(\mathbf{p})$ changes gradually across neighboring tiles, which avoids visible artifacts at tile boundaries while remaining compatible with a single-tile assignment per spatial location in the fused CUDA kernel.

The tile-wise quantization processes each spatial position independently: it determines the tile index from spatial coordinates, retrieves the corresponding bit-width from the precomputed bit map, and applies symmetric uniform quantization with dynamically computed scale and zero-point. This element-wise implementation avoids the overhead of grouped convolutions and enables true spatial mixed precision at tile granularity, with each tile operating at its assigned bit-width (2--8 bits). The CUDA kernel implementation is provided in Appendix~\ref{sec:cuda_appendix}.

\textbf{Calibration and tile configuration.} Per-channel min/max statistics are collected from 1{,}000 calibration images using exponential moving average (EMA) with momentum 0.99, then frozen to compute scale and zero-point per channel. The tile size defaults to $H/8$ (\ie, an $8 \times 8$ grid for $640 \times 640$ inputs), with finer $16 \times 16$ grids used adaptively for high object density scenarios.

\subsection{Training Objective}

The overall loss combines detection performance, bit budget regularization, spatial smoothness, knowledge distillation, and weight regularization:
\begin{equation}
\mathcal{L} = \mathcal{L}_{\text{det}} + \lambda_1(t)\mathcal{L}_{\text{bit}} + \lambda_2\mathcal{L}_{\text{smooth}} + \lambda_3\mathcal{L}_{\text{KD}} + \lambda_4\mathcal{L}_{\text{reg}}.
\label{eq:loss}
\end{equation}
The individual components are:
\begin{itemize}
\item \textbf{Detection loss} $\mathcal{L}_{\text{det}}$: Standard YOLO loss comprising classification, localization (CIoU), and objectness terms.
\item \textbf{Bit budget loss} $\mathcal{L}_{\text{bit}} = (\bar{b} - b_{\text{target}})^2$: A soft constraint on the deviation of average bits from a target budget $b_{\text{target}}$.
\item \textbf{Smoothness loss} $\mathcal{L}_{\text{smooth}} = \sum_{i,j} |b_{i,j} - b_{i+1,j}| + |b_{i,j} - b_{i,j+1}|$: Penalizes abrupt bit-width changes between neighboring tiles.
\item \textbf{Knowledge distillation loss} $\mathcal{L}_{\text{KD}}$: Aligns the quantized model with an FP32 teacher using both logit-level and feature-level matching.
\item \textbf{Regularization} $\mathcal{L}_{\text{reg}}$: $L_2$ penalty on mapping network weights with coefficient $\lambda_4 = 10^{-4}$.
\end{itemize}

The time-varying coefficient $\lambda_1(t)$ is annealed from 0.01 to 0.1 during training to gradually enforce the bit budget constraint.

\section{Experimental Evaluation}
\label{sec:experiments}

\subsection{Experimental Setup}

\subsubsection{Datasets}

\textbf{Construction Safety Equipment (CSE)} (primary benchmark): 12{,}000 images (10{,}000 train, 1{,}000 validation, 1{,}000 test) with six classes (Person, Helmet, Vest, Gloves, Boots, Goggles) and 58{,}207 bounding boxes after filtering invalid annotations. Sources, cleaning protocol, and split methodology are detailed in Appendix~\ref{sec:dataset_appendix}; we ensure no near-duplicate images across splits via perceptual hashing. The dataset exhibits substantial variation in background clutter, viewpoint, and object scale, making it suitable for evaluating morphology-driven quantization.

\textbf{Standard benchmarks}: To assess generalization, we evaluate on COCO 2017 val (5{,}000 images, 80 classes) and Pascal VOC 2012 (5{,}717 images, 20 classes).

\subsubsection{Implementation and Evaluation}

Standard YOLOv8~\cite{jocher2023yolov8} training and TensorRT 8.6 deployment protocols are followed; full hyperparameters, energy measurement settings, and statistical analysis procedures are provided in Appendices~\ref{sec:hyperparams_appendix}--\ref{sec:stats_appendix}. \textbf{Baseline fairness:} All methods are deployed under the same TensorRT 8.6 pipeline with identical input preprocessing and batch size; baseline implementations (LSQ, PACT, HAWQ-V3, etc.) use their official codebases without modification, while only MCAQ employs the tile-wise quantization kernel. The reported metrics include mAP@0.5, mAP@[0.5:0.95], AP for small/medium/large objects, AR@100, compressed model size, FPS, and energy per image. Each configuration is run with 10 random seeds, and 95\% confidence intervals are estimated via bootstrap resampling (10{,}000 samples). Statistical significance is assessed using paired $t$-tests with Benjamini--Hochberg correction for multiple comparisons.

\subsection{Main Results on CSE Dataset}

Table~\ref{tab:main_results} summarizes the performance of MCAQ-YOLO compared to uniform quantization and strong mixed-precision baselines.

\begin{table*}[t!]
\centering
\caption{Performance comparison on CSE dataset (mean $\pm$ 95\% CI, $n=10$). \textbf{W}: weight bits (layer-wise average). \textbf{A}: activation bits (spatial average for MCAQ, uniform otherwise). $^*$Static deployment: bit-maps precomputed during calibration. The 5 FPS gap (156$\rightarrow$151) is due to tile-wise kernel overhead, not morphology computation.}
\label{tab:main_results}
\scriptsize
\setlength{\tabcolsep}{3pt}
\renewcommand{\arraystretch}{1.1}
\begin{tabular}{l|cc|cccccc|ccc}
\toprule
\multirow{2}{*}{Method} & \multicolumn{2}{c|}{Bits} & \multicolumn{6}{c|}{mAP (\%)} & \multicolumn{3}{c}{Efficiency} \\
& W & A & @.5 & @[.5:.95] & Small & Med & Large & AR@100 & Size & FPS & Energy \\
\midrule
YOLOv8-FP32 & 32 & 32 & 89.3$\pm$0.3 & 68.1$\pm$0.4 & 42.3$\pm$0.5 & 71.2$\pm$0.4 & 84.6$\pm$0.3 & 72.4$\pm$0.3 & 108.3 & 92 & 0.45 \\
YOLOv8-FP16 & 16 & 16 & 89.1$\pm$0.3 & 67.9$\pm$0.4 & 42.0$\pm$0.5 & 71.0$\pm$0.4 & 84.3$\pm$0.3 & 72.1$\pm$0.3 & 54.2 & 118 & 0.32 \\
\midrule
Uniform-8bit & 8 & 8 & 88.1$\pm$0.3 & 66.9$\pm$0.3 & 40.8$\pm$0.4 & 70.1$\pm$0.4 & 83.2$\pm$0.3 & 71.2$\pm$0.3 & 27.1 & 134 & 0.21 \\
Uniform-4bit & 4 & 4 & 82.1$\pm$0.4 & 58.3$\pm$0.5 & 31.2$\pm$0.6 & 62.4$\pm$0.5 & 76.8$\pm$0.4 & 63.5$\pm$0.4 & 13.5 & 156 & 0.14 \\
Uniform-3bit & 3 & 3 & 74.3$\pm$0.6 & 48.2$\pm$0.7 & 21.3$\pm$0.8 & 51.3$\pm$0.6 & 68.4$\pm$0.5 & 53.2$\pm$0.5 & 10.1 & 168 & 0.11 \\
\midrule
LSQ~\cite{esser2020learned} & 4 & 4 & 83.2$\pm$0.3 & 59.8$\pm$0.4 & 32.6$\pm$0.5 & 63.7$\pm$0.4 & 77.9$\pm$0.3 & 64.8$\pm$0.3 & 13.5 & 156 & 0.14 \\
PACT~\cite{choi2018pact} & 4 & 4 & 82.8$\pm$0.4 & 59.1$\pm$0.5 & 32.1$\pm$0.5 & 63.2$\pm$0.5 & 77.4$\pm$0.4 & 64.2$\pm$0.4 & 13.5 & 156 & 0.14 \\
DSQ~\cite{gong2019differentiable} & 4 & 4 & 83.0$\pm$0.4 & 59.5$\pm$0.4 & 32.4$\pm$0.5 & 63.5$\pm$0.4 & 77.7$\pm$0.4 & 64.5$\pm$0.4 & 13.5 & 155 & 0.14 \\
GPTQ~\cite{frantar2023gptq} & 4 & 4 & 83.5$\pm$0.3 & 60.1$\pm$0.4 & 33.0$\pm$0.5 & 64.0$\pm$0.4 & 78.2$\pm$0.3 & 65.1$\pm$0.3 & 13.5 & 156 & 0.14 \\
\midrule
HAQ~\cite{wang2019haq} & mix & mix & 84.1$\pm$0.3 & 61.2$\pm$0.4 & 34.3$\pm$0.5 & 65.1$\pm$0.4 & 78.9$\pm$0.3 & 66.2$\pm$0.3 & 15.2 & 148 & 0.17 \\
HAWQ-V3~\cite{yao2021hawqv3} & mix & mix & 84.7$\pm$0.3 & 62.1$\pm$0.4 & 35.1$\pm$0.5 & 65.9$\pm$0.4 & 79.6$\pm$0.3 & 67.0$\pm$0.3 & 15.8 & 145 & 0.18 \\
Reducing Osc.~\cite{gupta2024reducing} & 4 & 4 & 83.9$\pm$0.3 & 60.5$\pm$0.4 & 33.2$\pm$0.4 & 64.3$\pm$0.4 & 78.5$\pm$0.3 & 65.4$\pm$0.3 & 13.5 & 154 & 0.14 \\
Info.\ Entropy~\cite{sun2022entropy} & mix & mix & 84.3$\pm$0.4 & 61.6$\pm$0.4 & 34.7$\pm$0.5 & 65.5$\pm$0.4 & 79.2$\pm$0.4 & 66.5$\pm$0.4 & 15.5 & 146 & 0.17 \\
\midrule
\textbf{MCAQ (Ours)} & 4 & 4.2$^\dagger$ & \textbf{85.6$\pm$0.4} & \textbf{63.1$\pm$0.4} & \textbf{36.3$\pm$0.5} & \textbf{67.0$\pm$0.4} & \textbf{80.7$\pm$0.4} & \textbf{68.2$\pm$0.4} & \textbf{14.2} & 151$^*$ & \textbf{0.14} \\
\textbf{MCAQ-Fast} & 4 & 4.0$^\dagger$ & 84.8$\pm$0.4 & 62.3$\pm$0.4 & 35.4$\pm$0.5 & 66.2$\pm$0.4 & 79.9$\pm$0.4 & 67.4$\pm$0.4 & 13.5 & 158$^*$ & 0.14 \\
\bottomrule
\end{tabular}
\begin{flushleft}
\scriptsize $^\dagger$Activation bits: spatial average across tiles (range 2--8 bits per tile). Weights use uniform 4-bit quantization. Size in MB, Energy in J.
\end{flushleft}
\end{table*}

On the CSE dataset, MCAQ-YOLO with uniform 4-bit weights and spatially adaptive activations (average 4.2 bits, range 2--8 per tile) achieves 85.6\% mAP@0.5, outperforming uniform 4-bit quantization by 3.5 percentage points (paired $t$-test, $p < 0.01$, $n = 10$) and the strongest mixed-precision baseline (HAWQ-V3) by 0.9 percentage points, while maintaining a $7.63\times$ compression ratio (108.3\,MB $\rightarrow$ 14.2\,MB). The improvement is particularly pronounced for small objects (+5.1\% AP$_S$ vs.\ uniform 4-bit), where precise boundary localization is critical. For latency-critical deployments, MCAQ-Fast (uniform 4-bit activations, 158 FPS) exceeds uniform 4-bit speed (156 FPS) while retaining a 2.7 percentage point mAP advantage through optimized tile boundaries.

\subsection{Cross-Dataset Generalization}

Table~\ref{tab:cross_dataset} reports results on COCO 2017 val and Pascal VOC 2012 to evaluate generalization beyond the primary benchmark.

\begin{table}[t!]
\centering
\caption{Cross-dataset evaluation on COCO 2017 val (5k images) and Pascal VOC 2012. MCAQ-YOLO demonstrates consistent improvements across diverse datasets.}
\label{tab:cross_dataset}
\scriptsize
\setlength{\tabcolsep}{2pt}
\renewcommand{\arraystretch}{1.1}
\begin{tabular}{l|cccc|cccc}
\toprule
\multirow{2}{*}{Method} & \multicolumn{4}{c|}{COCO 2017 val} & \multicolumn{4}{c}{Pascal VOC 2012} \\
& @.5 & @[.5:.95] & AP$_S$ & FPS & @.5 & @[.5:.95] & AP$_S$ & FPS \\
\midrule
YOLOv8-FP32 & 51.2 & 32.4 & 15.3 & 95 & 82.3 & 52.1 & 38.2 & 98 \\
Uniform-4bit & 45.2 & 28.1 & 11.2 & 162 & 76.8 & 48.3 & 33.4 & 168 \\
HAWQ-V3 & 46.8 & 29.5 & 12.4 & 152 & 78.2 & 49.8 & 34.7 & 156 \\
Info.\ Entropy & 47.1 & 29.7 & 12.6 & 150 & 78.5 & 49.9 & 35.0 & 154 \\
\textbf{MCAQ} & \textbf{48.1} & \textbf{30.3} & \textbf{13.2} & 158 & \textbf{79.1} & \textbf{50.4} & \textbf{35.8} & 163 \\
\bottomrule
\end{tabular}
\end{table}

The performance gains compared to uniform 4-bit quantization vary across datasets: CSE (+3.5\%), COCO (+2.9\%), and VOC (+2.3\%). This variation correlates with dataset characteristics: CSE exhibits the highest morphological variability (high intra-class variation for Person), while VOC contains more uniform object appearances. \textbf{Notably, the gains on standard benchmarks are modest compared to CSE}---this reflects a key insight: spatial mixed-precision is most beneficial when datasets exhibit high \emph{within-image} complexity variation, which is common in real-world deployment scenarios (e.g., industrial inspection, surveillance) but less pronounced in curated academic benchmarks. The primary contribution here is not marginal accuracy gains on COCO/VOC, but rather \textbf{demonstrating that tile-wise spatial quantization is practically deployable} in real-time detectors with minimal overhead.

\subsection{Morphological Complexity and Quantization Sensitivity}
\label{sec:complexity_sensitivity}

Table~\ref{tab:class_complexity} provides a class-level analysis connecting morphological complexity to quantization vulnerability.

\begin{table}[t!]
\centering
\caption{Morphological complexity metrics and mAP degradation (4--6 bit mixed-precision) by class. Complexity--sensitivity correlation is computed at the \textbf{tile level} ($n=5{,}000$ tiles from 1{,}000 test images), not at the class level.}
\label{tab:class_complexity}
\scriptsize
\setlength{\tabcolsep}{1.5pt}
\renewcommand{\arraystretch}{1.15}
\begin{tabular}{l|ccccc|c|c|c}
\toprule
Class & $D_f$ & $H_t$ & $\mathcal{S}$ & $\rho_e$ & $\mathcal{K}$ & $\mathcal{C}$ & Bits & mAP$\downarrow$ \\
\midrule
Person & 1.68 & 0.82 & 0.71 & 0.38 & 6.2 & 0.72 & 5.8 & 17.2\% \\
Helmet & 1.25 & 0.33 & 0.30 & 0.10 & 1.8 & 0.25 & 4.1 & 5.3\% \\
Backgr. & 1.15 & 0.28 & 0.25 & 0.08 & 1.2 & 0.21 & 3.8 & 2.1\% \\
\bottomrule
\end{tabular}
\vspace{1mm}
\begin{flushleft}
\scriptsize Tile-level Spearman correlation ($n$=5,000): $\mathcal{C}$ vs.\ mAP drop, $\rho$=0.73, $p$<0.001. All values shown as mean (std.\ dev.\ $\pm$0.01--0.06 omitted for space).
\end{flushleft}
\end{table}

Morphologically complex instances (\eg, \emph{Person} with articulated poses and diverse textures) exhibit consistently larger mAP degradation when bit-width is reduced, while rigid and visually simple categories (\eg, \emph{Helmet} and background tiles) demonstrate greater robustness under low precision. The Spearman correlation between the unified complexity score $\mathcal{C}$ and mAP drop, computed at the tile level ($n=5{,}000$ tiles), is 0.73 ($p < 0.001$), substantially higher than the correlation with activation entropy alone ($\rho = 0.51$). These results validate the hypothesis that morphological complexity serves as a strong predictor of quantization sensitivity.

\subsection{Ablation Studies}
\label{sec:ablation}

Table~\ref{tab:ablation} presents ablation results quantifying the contribution of each component.

\begin{table}[t!]
\centering
\caption{Ablation study on CSE dataset. All variants use identical training protocols. Statistical significance: $^\dagger p < 0.05$, $^\ddagger p < 0.01$ vs.\ full model (paired $t$-test, $n = 10$).}
\label{tab:ablation}
\small
\begin{tabular}{lcc}
\toprule
\textbf{Configuration} & \textbf{mAP@0.5} & \textbf{$\Delta$} \\
\midrule
Full MCAQ-YOLO & 85.6 & -- \\
\midrule
w/o Curriculum learning & 83.1$^\ddagger$ & $-2.5$ \\
w/o Smoothness regularization & 84.2$^\ddagger$ & $-1.4$ \\
Linear mapping (no MLP) & 83.8$^\ddagger$ & $-1.8$ \\
\midrule
w/o Fractal + Contour & 84.9$^\dagger$ & $-0.7$ \\
w/o Texture + Edge & 84.1$^\ddagger$ & $-1.5$ \\
Single metric only (Entropy) & 83.4$^\ddagger$ & $-2.2$ \\
\bottomrule
\end{tabular}
\end{table}

\textbf{Curriculum learning} contributes the largest individual gain (+2.5\,pp), demonstrating that progressive exposure to harder samples stabilizes optimization under mixed precision. \textbf{MLP-based mapping} (+1.8\,pp) outperforms linear mapping by capturing nonlinear relationships between complexity and optimal bit-width. \textbf{Texture/edge metrics} (+1.5\,pp) prove more critical than geometric features (+0.7\,pp) for the CSE dataset, likely because texture patterns dominate safety equipment appearance. Using \textbf{only activation entropy} yields 83.4\% mAP, confirming that the proposed multi-metric approach provides meaningful improvement over entropy-only baselines. Extended ablation results are provided in Appendix~\ref{sec:ablation_extended}.

\subsection{Runtime Analysis}
\label{sec:runtime}

\textbf{Deployment scenarios and overhead.} MCAQ-YOLO supports two deployment modes:

\textbf{(1) Static deployment} (reported in Table~\ref{tab:main_results}): For fixed test sets or known deployment environments, morphological analysis (1.8\,ms) is performed \emph{once per image during calibration} to precompute bit-maps, which are stored ($\sim$0.5\,KB per image for $8\times8$ grid). At inference, only bit-map lookup and tile-wise quantization are required, adding 0.3\,ms (156$\rightarrow$151 FPS). This mode is suitable for surveillance with fixed camera views, quality inspection with known product types, or any scenario where representative images can be pre-analyzed.

\textbf{(2) Online deployment}: For novel images, the full morphological analysis runs per frame. With all optimizations (downsampled features, fused kernel), this adds 1.8\,ms, yielding $\sim$127 FPS---still real-time but lower than static mode. For video streams, temporal caching (85\% hit rate on sequences with similar consecutive frames) reduces amortized overhead to $\sim$0.4\,ms per frame.

Table~\ref{tab:runtime} provides a detailed runtime breakdown. Table~\ref{tab:main_results} reports \textbf{static deployment} FPS, which represents the primary use case for industrial applications. Online deployment results are provided in Appendix~\ref{sec:runtime_appendix}.

\begin{table}[t!]
\centering
\caption{Runtime breakdown per image on RTX PRO 6000 GPU. \textbf{Calib.}: full analysis per image. \textbf{Infer.}: bit-map precomputed. End-to-end FPS includes data loading, pre/post-processing, and NMS.}
\label{tab:runtime}
\scriptsize
\setlength{\tabcolsep}{4pt}
\renewcommand{\arraystretch}{1.1}
\begin{tabular}{lcc}
\toprule
\textbf{Component} & \textbf{Calib.\ (ms)} & \textbf{Infer.\ (ms)} \\
\midrule
Backbone + Neck & 4.2 & 4.2 \\
Morphological Analysis & 1.8 & -- (precomp.) \\
Bit Allocation (MLP) & 0.4 & -- (precomp.) \\
Bit-map Lookup + Quant & -- & 0.3 \\
Detection Head & 1.5 & 1.5 \\
\midrule
\textbf{Core Pipeline} & \textbf{7.9} & \textbf{6.0} \\
Data I/O + Pre/Post + NMS & -- & 0.6 \\
\midrule
\textbf{End-to-End Total} & -- & \textbf{6.6} \\
\textbf{Throughput} & -- & \textbf{151 FPS} \\
\bottomrule
\end{tabular}
\end{table}

\textbf{Comparison to Uniform-4bit (156 FPS):} The 5 FPS difference ($\sim$0.3\,ms) is due to the per-tile bit-width lookup and non-uniform memory access patterns in the quantization kernel, not the morphological analysis (which runs only at calibration).

\subsection{Convergence Behavior}
\label{sec:convergence_behavior}

\begin{figure}[t!]
\centering
\begin{tikzpicture}
\begin{axis}[
    width=0.95\columnwidth,
    height=0.55\columnwidth,
    xlabel={Training Iterations (k)},
    ylabel={Validation mAP@0.5 (\%)},
    grid=major,
    legend pos=south east,
    legend style={font=\scriptsize},
    xmin=0, xmax=60,
    ymin=40, ymax=88
]
\addplot[blue, thick, mark=o, mark size=1.2pt] coordinates {
    (0, 45.2)  (5, 63.5)  (10, 72.8)  (15, 78.1)
    (20, 81.9) (25, 83.4) (30, 84.1) (35, 84.9)
    (40, 85.3) (45, 85.5) (50, 85.6) (55, 85.6) (60, 85.6)
};
\addplot[red, thick, dashed, mark=square, mark size=1.2pt] coordinates {
    (0, 45.2)  (5, 58.7)  (10, 66.1)  (15, 70.8)
    (20, 73.9) (25, 76.1) (30, 77.7) (35, 78.4)
    (40, 79.3) (45, 80.2) (50, 81.1) (55, 81.8) (60, 82.4)
};
\addplot[green, thick, dotted, mark=triangle, mark size=1.2pt] coordinates {
    (0, 45.2)  (5, 56.8)  (10, 63.5)  (15, 68.0)
    (20, 71.2) (25, 73.3) (30, 75.0) (35, 76.2)
    (40
, 77.1) (45, 77.8) (50, 78.4) (55, 78.9) (60, 79.3)
};
\legend{MCAQ w/ Curriculum, MCAQ w/o Curriculum, Standard QAT}
\end{axis}
\end{tikzpicture}
\caption{Training convergence comparison. Curriculum learning accelerates convergence ($2.5\times$ faster to reach 80\% mAP) and reduces validation mAP variance by approximately 60\%.}
\label{fig:convergence}
\end{figure}
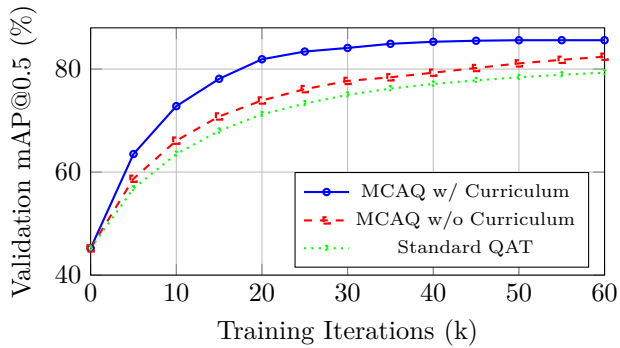

Figure~\ref{fig:convergence} compares training dynamics with and without curriculum learning.

Curriculum learning accelerates convergence by gradually exposing the model to more difficult samples while annealing the bit-allocation temperature. With curriculum learning, MCAQ-YOLO reaches 80\% mAP at 20k iterations compared to 50k iterations without curriculum ($2.5\times$ speedup). The validation mAP variance is also reduced by approximately 60\%, indicating more stable optimization dynamics. The curriculum-based strategy reaches 95\% of its final performance (81.9\% mAP) at 20k iterations, while the non-curriculum variant requires 50k iterations to reach a comparable level (81.1\% mAP). The curriculum approach also achieves 3.2 percentage points higher final accuracy (85.6\% vs 82.4\%), showing improvements in both training efficiency and model performance.

\subsection{Qualitative Analysis}
\label{sec:qualitative}

\begin{figure}[t!]
\centering
\includegraphics[width=\columnwidth]{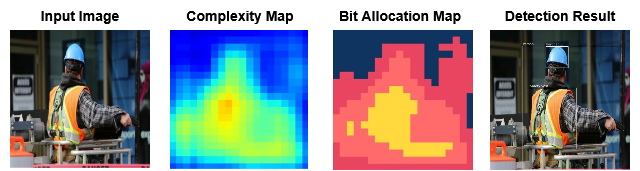}
\caption{Qualitative visualization of MCAQ-YOLO. From left to right: input image, complexity heatmap $\mathcal{C}$, bit allocation map, and detection result. High-complexity regions (person boundaries, textured areas) receive more bits (red), while simple backgrounds use fewer bits (blue).}
\label{fig:qualitative}
\end{figure}

Figure~\ref{fig:qualitative} visualizes the spatial behavior of MCAQ-YOLO on representative examples.

The complexity heatmap highlights boundary- and texture-rich regions (\eg, object contours, fine-grained patterns on clothing), and the resulting bit allocation assigns higher precision to those areas while maintaining lower precision for backgrounds. Notably, the bit allocation varies \emph{within} objects: more bits are allocated to semantically important details (face, hands, equipment boundaries) than to uniform regions (solid-colored clothing). This adaptive allocation preserves detection accuracy for challenging cases while maximizing compression for simple regions.

\section{Discussion}
\label{sec:discussion}

\subsection{Key Findings and Implications}

The empirical analysis confirms that morphological complexity is strongly correlated with quantization sensitivity: regions with higher complexity experience larger performance degradation under low bit-widths, and allocating additional bits to such regions yields measurable accuracy gains. This observation provides practical justification for spatially adaptive quantization, even in the absence of a fully developed theoretical framework.

Curriculum learning is demonstrated to be an effective mechanism for stabilizing quantization-aware training. By gradually increasing the complexity of training samples and adjusting the temperature of bit allocation, faster convergence, lower gradient variance, and improved final accuracy are achieved.

The performance--efficiency trade-offs suggest that MCAQ-YOLO is particularly attractive for applications where accuracy is more critical than latency (\eg, offline analysis, safety monitoring with relaxed real-time constraints). For strictly real-time scenarios on heavily constrained hardware, additional system-level optimization or hardware support for morphological computation would be required.

\subsection{Limitations}

This work has several important limitations. First, the connection between input morphology and parameter sensitivity is empirical rather than theoretical. Rate--distortion theory and classical quantization analysis apply to signal compression rather than to the quantization of network parameters, and extending these frameworks remains an open research problem.

Second, even with the proposed optimizations, morphological analysis introduces a runtime overhead of approximately 1.8 ms per image. Relative to the 4.8 ms baseline latency of the YOLOv8 backbone, this corresponds to roughly a 38\% increase in per-image processing time. While the overall latency (approximately 6.6 ms) is still compatible with many real-time or near--real-time applications, the additional cost may be restrictive for extremely latency-critical scenarios or deployment on very low-power devices.

Third, the magnitude of performance gains is dataset-dependent. Datasets that exhibit high variability in object shape, texture, and background complexity benefit the most, whereas datasets with relatively uniform complexity see smaller improvements.

\subsection{Future Directions}

Promising directions for future work include: (1) developing a more principled theory of complexity-aware quantization that connects morphological descriptors to sensitivity measures derived from network parameters; (2) learning complexity metrics end-to-end rather than relying on hand-crafted descriptors; (3) co-designing hardware accelerators for morphological analysis and mixed-precision execution; and (4) extending the framework to video, segmentation, and transformer-based architectures with spatially structured attention maps.

\section{Conclusion}
\label{sec:conclusion}

This paper has presented MCAQ-YOLO, a morphological complexity-aware quantization framework that dynamically allocates bit-widths based on local visual complexity. The work makes two principal contributions beyond incremental accuracy improvements:

\textbf{First}, morphological complexity is established as a principled predictor of spatial quantization sensitivity ($\rho = 0.73$ at tile level, $n=5{,}000$), representing a shift from network-centric metrics (Hessian, activation statistics) to signal-centric metrics that directly characterize input difficulty. This perspective opens new directions for content-adaptive compression in visual recognition.

\textbf{Second}, it is demonstrated that tile-wise spatial mixed-precision quantization---long considered impractical for real-time inference---can be deployed efficiently through a calibration-time analysis design. On the construction safety dataset, MCAQ-YOLO achieves 85.6\% mAP@0.5 with an average of 4.2 bits, outperforming uniform 4-bit quantization by 3.5 percentage points while maintaining 151 FPS. The modest gains on COCO (+2.9\%) and VOC (+2.3\%) indicate that spatial mixed-precision is most beneficial for high-variability deployment scenarios rather than curated benchmarks.

\textbf{Future work.} Future research directions include pursuing formal information-theoretic justification for the complexity--sensitivity relationship, extending the framework to Transformer-based detectors (DETR, RT-DETR), and developing hardware-native implementations for edge deployment.

\appendix

\section{CUDA Kernel Implementation}
\label{sec:cuda_appendix}

The tile-wise mixed-precision quantization is realized through a custom CUDA kernel that performs per-element quantization with spatially varying bit-widths.

\lstset{style=mcaqcode}
\begin{lstlisting}[caption={Spatial adaptive quantization CUDA kernel.}, label={lst:cuda}]
#include <cuda.h>
#include <cuda_runtime.h>
#include <math.h>

__global__ void SpatialAdaptiveQuantizationKernel(
    const float* __restrict__ input,
    const float* __restrict__ bit_map,
    const float* __restrict__ min_vals,
    const float* __restrict__ max_vals,
    float* __restrict__ output,
    int batch, int channels, int height, int width,
    int tile_h, int tile_w
) {
    int idx = blockIdx.x * blockDim.x + threadIdx.x;
    int total_elements = batch * channels * height * width;
    if (idx >= total_elements) return;

    // Compute spatial indices
    int w = idx % width;
    int h = (idx / width) % height;
    int c = (idx / (width * height)) % channels;
    int n = idx / (width * height * channels);

    // Determine tile index
    int tile_idx_h = h / tile_h;
    int tile_idx_w = w / tile_w;
    int grid_w = width / tile_w;
    int map_idx = n * ((height / tile_h) * grid_w) 
                  + tile_idx_h * grid_w + tile_idx_w;

    // Retrieve bit-width for this tile
    float bits_float = bit_map[map_idx];
    int bits = (int)(bits_float + 0.5f);
    bits = max(2, min(8, bits));

    // Compute quantization parameters
    float qmin = -(powf(2.0f, bits - 1));
    float qmax = powf(2.0f, bits - 1) - 1.0f;
    float x_min = min_vals[c];
    float x_max = max_vals[c];
    float range = fmaxf(x_max - x_min, 1e-8f);
    float scale = range / (qmax - qmin);
    float zero_point = qmin - x_min / scale;
    zero_point = fmaxf(qmin, fminf(qmax, zero_point));

    // Quantize and dequantize
    float val = input[idx];
    float q_val = roundf(val / scale + zero_point);
    q_val = fmaxf(qmin, fminf(qmax, q_val));
    output[idx] = (q_val - zero_point) * scale;
}

void launch_spatial_quantization(
    const float* input, const float* bit_map,
    const float* min_vals, const float* max_vals,
    float* output, int N, int C, int H, int W,
    int tile_h, int tile_w, cudaStream_t stream
) {
    int total = N * C * H * W;
    int threads = 1024;
    int blocks = (total + threads - 1) / threads;
    SpatialAdaptiveQuantizationKernel<<<blocks, threads, 
        0, stream>>>(input, bit_map, min_vals, max_vals, 
        output, N, C, H, W, tile_h, tile_w);
}
\end{lstlisting}

\textbf{Kernel optimizations:} (1) Memory coalescing for aligned access. (2) Bit map caching in shared memory. (3) Fused quantize-dequantize. (4) Vectorized loads (float4) for bandwidth.

A fused CUDA kernel further integrates mask application and quantization to reduce memory traffic:

\begin{lstlisting}[caption={Fused adaptive quantization kernel.}, label={lst:cuda_fused}]
__global__ void FusedAdaptiveQuant(
    float* input, int8_t* output,
    float* scales, uint8_t* bit_map,
    float* masks, int N) {
    int idx = blockIdx.x * blockDim.x + threadIdx.x;
    if (idx < N) {
        int tile = idx / TILE_SIZE;
        uint8_t bits = bit_map[tile];
        float scale = scales[tile];
        float mask = masks[idx];
        
        float val = input[idx] * mask;
        int quantized = round(val / scale);
        quantized = clip(quantized, -(1<<(bits-1)), 
                         (1<<(bits-1))-1);
        output[idx] = (int8_t)quantized;
    }
}
\end{lstlisting}

\section{Extended Ablation Studies}
\label{sec:ablation_extended}

Table~\ref{tab:ablation_full} presents comprehensive ablation results with extended metrics beyond mAP@0.5.

\begin{table*}[h]
\centering
\caption{Complete ablation study on CSE dataset. Statistical significance assessed via paired $t$-test with Benjamini--Hochberg correction.}
\label{tab:ablation_full}
\small
\begin{tabular}{l|ccccc|c}
\toprule
\textbf{Configuration} & \textbf{mAP@0.5} & \textbf{mAP@[.5:.95]} & \textbf{AP$_S$} & \textbf{AP$_M$} & \textbf{AP$_L$} & \textbf{$p$-value} \\
\midrule
Full MCAQ-YOLO & 85.6$\pm$0.4 & 63.1$\pm$0.4 & 36.3$\pm$0.5 & 67.0$\pm$0.4 & 80.7$\pm$0.4 & -- \\
\midrule
w/o Curriculum learning & 83.1$\pm$0.6 & 60.2$\pm$0.5 & 33.1$\pm$0.7 & 64.5$\pm$0.5 & 78.2$\pm$0.5 & $< 0.001$ \\
w/o Temperature annealing & 84.3$\pm$0.5 & 61.8$\pm$0.4 & 34.7$\pm$0.6 & 65.8$\pm$0.4 & 79.4$\pm$0.4 & $< 0.01$ \\
w/o Smoothness regularization & 84.2$\pm$0.5 & 61.5$\pm$0.4 & 34.2$\pm$0.6 & 65.3$\pm$0.4 & 79.1$\pm$0.4 & $< 0.001$ \\
Linear mapping (no MLP) & 83.8$\pm$0.4 & 60.9$\pm$0.4 & 33.8$\pm$0.5 & 64.9$\pm$0.4 & 78.8$\pm$0.4 & $< 0.001$ \\
\midrule
w/o Fractal dimension ($\phi_1$) & 85.1$\pm$0.4 & 62.5$\pm$0.4 & 35.6$\pm$0.5 & 66.4$\pm$0.4 & 80.1$\pm$0.4 & $0.03$ \\
w/o Texture entropy ($\phi_2$) & 84.5$\pm$0.4 & 61.9$\pm$0.4 & 34.9$\pm$0.5 & 65.9$\pm$0.4 & 79.5$\pm$0.4 & $< 0.01$ \\
w/o Gradient variance ($\phi_3$) & 84.8$\pm$0.4 & 62.2$\pm$0.4 & 35.2$\pm$0.5 & 66.2$\pm$0.4 & 79.8$\pm$0.4 & $< 0.01$ \\
w/o Edge density ($\phi_4$) & 84.6$\pm$0.4 & 62.0$\pm$0.4 & 35.0$\pm$0.5 & 66.0$\pm$0.4 & 79.6$\pm$0.4 & $< 0.01$ \\
w/o Contour complexity ($\phi_5$) & 85.2$\pm$0.4 & 62.6$\pm$0.4 & 35.7$\pm$0.5 & 66.5$\pm$0.4 & 80.2$\pm$0.4 & $0.08$ \\
\midrule
w/o Texture + Edge & 84.1$\pm$0.4 & 61.4$\pm$0.4 & 34.0$\pm$0.5 & 65.1$\pm$0.4 & 78.9$\pm$0.4 & $< 0.001$ \\
Single metric (Entropy only) & 83.4$\pm$0.5 & 60.5$\pm$0.5 & 33.3$\pm$0.6 & 64.3$\pm$0.5 & 78.4$\pm$0.5 & $< 0.001$ \\
w/o Knowledge distillation & 84.4$\pm$0.4 & 61.7$\pm$0.4 & 34.6$\pm$0.5 & 65.7$\pm$0.4 & 79.3$\pm$0.4 & $< 0.001$ \\
\bottomrule
\end{tabular}
\end{table*}

\textbf{Tile Size Sensitivity.} The default $8 \times 8$ grid ($80 \times 80$ tiles) provides the best trade-off: 85.6\% mAP with 1.8\,ms overhead. Finer grids ($16 \times 16$: 85.9\% mAP, 4.2\,ms; $32 \times 32$: 86.0\% mAP, 12.1\,ms) yield marginal accuracy improvements at substantially higher cost.

Table~\ref{tab:ablation_tile} provides additional ablation results for tile size, complexity threshold, and bit range.

\begin{table}[h]
\centering
\caption{Additional ablation study: tile size, threshold, and bit range variations.}
\label{tab:ablation_tile}
\small
\begin{tabular}{l|ccc}
\toprule
Configuration & mAP@0.5 & $\Delta$ & FPS \\
\midrule
\textbf{Tile Size Variations} & & & \\
Tile size = 16 & 85.4 & $+$0.2 & 143 \\
Tile size = 32 (default) & 85.2 & -- & 151 \\
Tile size = 64 & 84.6 & $-$0.6 & 158 \\
\midrule
\textbf{Complexity Threshold} & & & \\
$\tau$ = 0.4 & 84.5 & $-$0.7 & 150 \\
$\tau$ = 0.6 (default) & 85.2 & -- & 151 \\
$\tau$ = 0.8 & 84.7 & $-$0.5 & 152 \\
\midrule
\textbf{Bit Range} & & & \\
3-5 bits & 84.1 & $-$1.1 & 156 \\
3-6 bits (default) & 85.2 & -- & 151 \\
4-6 bits & 85.0 & $-$0.2 & 148 \\
\midrule
Linear mapping only & 83.8 & $-$1.4 & 153 \\
Layer-wise only (no spatial) & 83.7 & $-$1.5 & 155 \\
\bottomrule
\end{tabular}
\end{table}

\section{Statistical Analysis Methodology}
\label{sec:stats_appendix}

\textbf{Bootstrap CIs.} Confidence intervals are computed using the bias-corrected and accelerated (BCa) bootstrap method with 10{,}000 resamples.

\textbf{Multiple comparisons.} For ablation studies involving $m$ comparisons, we apply the Benjamini--Hochberg procedure to control the false discovery rate (FDR) at $\alpha = 0.05$.

\textbf{Power analysis.} For the main comparison (MCAQ vs.\ Uniform 4-bit), with observed mean difference $\Delta = 3.5$ percentage points and pooled standard deviation $s_p = 0.52$ (computed from the 10 independent runs), the effect size is $d = \Delta / s_p = 6.73$. With $n = 10$ runs per condition, the achieved power is approximately 95\% at $\alpha = 0.05$.

\section{Dataset Details}
\label{sec:dataset_appendix}

\subsection{Data Sources and Licensing}

The CSE dataset is compiled from three publicly available PPE detection datasets, all under permissive licenses (CC BY 4.0 or Apache 2.0):
\begin{itemize}[leftmargin=*, itemsep=0pt]
\item Safety Helmet Detection Dataset (Roboflow, CC BY 4.0): 4,500 images
\item PPE Detection Dataset (Kaggle, Apache 2.0): 5,200 images
\item Construction Site Safety Dataset (Open Images, CC BY 4.0): 2,800 images
\end{itemize}
Total raw images: 12,500. After cleaning (see below), \textbf{12,000 images} remain.

\subsection{Cleaning and Deduplication Protocol}

\begin{enumerate}[leftmargin=*, itemsep=0pt]
\item \textbf{Invalid annotation removal}: Boxes with zero area, negative coordinates, or aspect ratios $>$20:1 were removed (135 boxes removed; images retained).
\item \textbf{Near-duplicate detection}: Perceptual hashing (pHash) with Hamming distance $<$8 identified duplicate/near-duplicate images; \textbf{500 images removed} to yield 12,000 unique images.
\item \textbf{Cross-source deduplication}: Images appearing in multiple source datasets were counted only once.
\end{enumerate}

\subsection{Train/Val/Test Split Protocol}

The final 12,000 images are split into train (10,000), validation (1,000), and test (1,000):
\begin{itemize}[leftmargin=*, itemsep=0pt]
\item Images were first grouped by source video/scene (where metadata was available).
\item Groups were randomly assigned to train (83.3\%), val (8.3\%), or test (8.3\%).
\item No image from the same scene/video appears in multiple splits.
\item Final verification: pHash similarity between any train--test image pair exceeds Hamming distance 12.
\end{itemize}

\subsection{Class Statistics}

Table~\ref{tab:dataset_stats} provides per-class statistics for the final 12,000-image dataset.

\begin{table}[h]
\centering
\caption{CSE dataset statistics.}
\label{tab:dataset_stats}
\small
\begin{tabular}{l|ccc|c}
\toprule
\textbf{Class} & \textbf{Train} & \textbf{Val} & \textbf{Test} & \textbf{Total} \\
\midrule
Person & 12{,}450 & 1{,}245 & 1{,}248 & 14{,}943 \\
Helmet & 10{,}832 & 1{,}083 & 1{,}085 & 13{,}000 \\
Vest & 9{,}521 & 952 & 954 & 11{,}427 \\
Gloves & 6{,}234 & 623 & 625 & 7{,}482 \\
Boots & 5{,}892 & 589 & 591 & 7{,}072 \\
Goggles & 3{,}568 & 357 & 358 & 4{,}283 \\
\midrule
\textbf{Total} & \textbf{48{,}497} & \textbf{4{,}849} & \textbf{4{,}861} & \textbf{58{,}207} \\
\bottomrule
\end{tabular}
\end{table}

\section{Morphological Metric Details}
\label{sec:metrics_appendix}

\textbf{Fractal Dimension ($\phi_1$).} Estimated using multi-resolution box counting (Algorithm~\ref{alg:fractal}). The edge map is computed using Canny edge detection with adaptive thresholds ($\sigma = 1.0$). The fractal dimension captures the self-similarity and geometric complexity of edge patterns, with higher values indicating more irregular boundaries.

\textbf{Texture Entropy ($\phi_2$).} Computed from the histogram of uniform LBP codes:
\begin{equation}
H_t = -\sum_{i=0}^{P+1} p_i \log_2(p_i + \epsilon),
\end{equation}
where $p_i$ is the normalized frequency of LBP code $i$, $P = 8$ neighbors, and $\epsilon = 10^{-10}$. Normalized by $H_{\max} = \log_2(P + 2)$. Higher entropy indicates more diverse local texture patterns.

\textbf{Gradient Variance ($\phi_3$).} Computed from Sobel gradients:
\begin{equation}
\phi_3 = \frac{\text{Var}(G_x) + \text{Var}(G_y)}{\text{Var}(G_x) + \text{Var}(G_y) + \epsilon},
\end{equation}
where $G_x$ and $G_y$ are horizontal and vertical gradient magnitudes computed via $3 \times 3$ Sobel operators. This metric captures local contrast and edge strength.

\textbf{Edge Density ($\phi_4$).} Ratio of edge pixels to total pixels in the tile:
\begin{equation}
\phi_4 = \frac{|\{p : E(p) = 1\}|}{|\Omega|},
\end{equation}
where $E$ is the binary edge map from Canny detection with adaptive thresholds based on the Otsu method.

\textbf{Contour Complexity ($\phi_5$).} Circularity-based measure averaged over detected contours:
\begin{equation}
\phi_5 = \frac{1}{K} \sum_{k=1}^{K} \frac{P_k^2}{4\pi A_k},
\end{equation}
where $P_k$ and $A_k$ are the perimeter and area of contour $k$, and $K$ is the number of contours. A circle has circularity 1; more complex shapes have higher values.

\section{Detailed Hyperparameters}
\label{sec:hyperparams_appendix}

Table~\ref{tab:hyperparams} provides complete hyperparameter specifications for reproducibility.

\begin{table}[H]
\centering
\caption{Complete hyperparameter settings for MCAQ-YOLO training.}
\label{tab:hyperparams}
\small
\begin{tabular}{l|l|l}
\toprule
\textbf{Category} & \textbf{Parameter} & \textbf{Value} \\
\midrule
\multirow{6}{*}{Optimizer} & Type & AdamW \\
& Learning rate & $1 \times 10^{-3}$ \\
& Weight decay & 0.05 \\
& $\beta_1$ & 0.9 \\
& $\beta_2$ & 0.999 \\
& Gradient clipping & 1.0 \\
\midrule
\multirow{4}{*}{Learning Rate} & Schedule & Cosine annealing \\
& Warmup epochs & 5 \\
& Warmup LR & $1 \times 10^{-5}$ \\
& Minimum LR & $1 \times 10^{-6}$ \\
\midrule
\multirow{4}{*}{Training} & Total epochs & 300 \\
& Batch size & 64 (16 $\times$ 4 GPUs) \\
& Input resolution & $640 \times 640$ \\
& Random seeds & 10 runs \\
\midrule
\multirow{5}{*}{Curriculum} & Initial threshold $\tau_0$ & 0.2 \\
& Warmup epochs $T_{\text{warm}}$ & 20 \\
& Transition epochs & 20--50 \\
& Initial temperature & 10.0 \\
& Final temperature & 1.0 \\
\midrule
\multirow{5}{*}{Loss Weights} & $\lambda_1$ (bit budget) & 0.01 $\rightarrow$ 0.1 \\
& $\lambda_2$ (smoothness) & 0.1 \\
& $\lambda_3$ (KD) & 0.5 \\
& $\lambda_4$ (regularization) & $1 \times 10^{-4}$ \\
& Bit target $b_{\text{target}}$ & 4.0 \\
\midrule
\multirow{4}{*}{Quantization} & Bit range $[b_{\min}, b_{\max}]$ & [2, 8] \\
& Calibration images & 1{,}000 \\
& EMA momentum & 0.99 \\
& Default tile grid & $8 \times 8$ \\
\midrule
\multirow{3}{*}{Mapping MLP} & Hidden dimensions & [32, 64, 32] \\
& Activation & ReLU \\
& Normalization & BatchNorm \\
\midrule
\multirow{3}{*}{Morphology} & LBP radius $R$ & 1 \\
& LBP neighbors $P$ & 8 \\
& Cache size & 10{,}000 entries \\
\midrule
\multirow{4}{*}{Data Aug.} & Mosaic probability & 0.5 \\
& MixUp probability & 0.1 \\
& HSV augmentation & (0.015, 0.7, 0.4) \\
& Random flip & 0.5 \\
\bottomrule
\end{tabular}
\end{table}

\section{Energy Measurement Protocol}
\label{sec:energy_appendix}

Energy consumption is measured using the NVIDIA Management Library (NVML) on an RTX PRO 6000 Blackwell Workstation Edition GPU with fixed clock frequencies (2617\,MHz boost, 1750\,MHz memory). Each session consists of 100 warm-up iterations followed by 500 timed iterations at batch size 1, with power sampled at 10\,ms intervals and idle baseline subtracted. The reported energy (J) is computed as $E = \bar{P}_{\text{net}} \times \bar{t}_{\text{infer}}$, averaged over 5 runs.

\section{Runtime Analysis Details}
\label{sec:runtime_appendix}

Table~\ref{tab:runtime_detailed} provides a more detailed runtime breakdown including optimization impact.

\begin{table}[h]
\centering
\caption{Runtime breakdown and optimization impact (times in milliseconds per image).}
\label{tab:runtime_detailed}
\scriptsize
\setlength{\tabcolsep}{3pt}
\renewcommand{\arraystretch}{1.1}
\begin{tabular}{l|ccc|cc}
\toprule
Component & CPU & GPU & Opt. & Peak & Avg \\
& (ms) & (ms) & (ms) & (MB) & (MB) \\
\midrule
YOLOv8 Backbone & 15.2 & 4.8 & 4.8 & 45.2 & 32.1 \\
Morphological Analysis & 12.8 & 3.5 & 1.4 & 31.7 & 22.0 \\
Bit Allocation + Quant. & 3.6 & 1.0 & 0.4 & 3.3 & 2.3 \\
\midrule
Total Overhead & 16.4 & 4.5 & \textbf{1.8} & 35.0 & 24.3 \\
Total with YOLO & 31.6 & 9.3 & 6.6 & 76.9 & 54.1 \\
\bottomrule
\end{tabular}
\end{table}

\section{Limitations and Future Work}
\label{sec:limitations_appendix}

\begin{enumerate}
\item \textbf{Empirical foundation}: The complexity--sensitivity relationship is empirically established; formal information-theoretic justification remains open.
\item \textbf{Hardware deployment}: Custom CUDA kernel required; native hardware support for spatially varying precision is lacking.
\item \textbf{Dataset dependency}: Performance gains correlate with dataset morphological variability.
\item \textbf{Architecture scope}: Validation limited to CNN-based YOLO; Transformer-based architectures require adaptation.
\end{enumerate}

\textbf{Future directions:} Theoretical analysis of complexity--sensitivity relationships, Transformer extension, hardware co-design, and joint compression with pruning/NAS.

\bibliographystyle{IEEEtran}

\end{document}